\documentclass[10pt,twocolumn,letterpaper]{article}

\usepackage{cvpr}
\usepackage{times}
\usepackage{epsfig}
\usepackage{graphicx}
\usepackage{amssymb}
\usepackage{times}  %Required
\usepackage{helvet}  %Required
\usepackage{courier}  %Required
\usepackage{url}  %Required
\usepackage{graphicx}  %Required
\usepackage{multirow,floatrow, comment, enumerate}
\usepackage{amsmath,amssymb}
\usepackage{color}
\usepackage{algorithmic}
\usepackage[linesnumbered,ruled,vlined]{algorithm2e}
\usepackage{subcaption}
\usepackage[breaklinks=true,bookmarks=false]{hyperref}

\cvprfinalcopy % *** Uncomment this line for the final submission

\def\cvprPaperID{1535} % *** Enter the CVPR Paper ID here

\begin{document}

%%%%%%%%% TITLE
\title{Detach and Adapt: Learning Cross-Domain Disentangled Deep Representation}

\author{Yen-Cheng Liu$^{1}$, Yu-Ying Yeh$^{1}$, Tzu-Chien Fu$^{2}$, Sheng-De Wang$^{1}$, \\
Wei-Chen Chiu$^{3}$, Yu-Chiang Frank Wang$^{1}$\\
$^1$Department of Electrical Engineering, National Taiwan University, Taiwan\\
$^2$Department of Electrical Engineering \& Computer Science, Northwestern University, USA\\
$^3$Department of Computer Science, National Chiao Tung University, Taiwan\\
{\tt\small \{r04921003, b99202023\}@ntu.edu.tw, tcfu@u.northwestern.edu, sdwang@ntu.edu.tw,}\\
{\tt\small walon@cs.nctu.edu.tw, ycwang@ntu.edu.tw}
% For a paper whose authors are all at the same institution,
% omit the following lines up until the closing ``}''.
% Additional authors and addresses can be added with ``\and'',
% just like the second author.
% To save space, use either the email address or home page, not both
}

\maketitle
%\thispagestyle{empty}

%%%%%%%%% ABSTRACT
\begin{abstract}
%%%%%%%%% ABSTRACT
While representation learning aims to derive interpretable features for describing visual data, representation disentanglement further results in such features so that particular image attributes can be identified and manipulated. However, one cannot easily address this task without observing ground truth annotation for the training data.
To address this problem, we propose a novel deep learning model of Cross-Domain Representation Disentangler (CDRD). By observing fully annotated source-domain data and unlabeled target-domain data of interest, our model bridges the information across data domains and transfers the attribute information accordingly. Thus, cross-domain feature disentanglement and adaptation can be jointly performed. In the experiments, we provide qualitative results to verify our disentanglement capability. Moreover, we further confirm that our model can be applied for solving classification tasks of unsupervised domain adaptation, and performs favorably against state-of-the-art image disentanglement and translation methods.
\end{abstract}

%%%%%%%%% BODY TEXT %%%%%%%%%%%
\section{Introduction} \label{sec:intro}
The development of deep neural networks benefits a variety of areas such as computer vision, machine learning, and natural language processing, which results in promising progresses in realizing artificial intelligence environments. However, as pointed out in~\cite{bengio2013representation}, it is fundamental and desirable for understanding the observed information around us. To be more precise, the above goal is achieved by identifying and disentangling the underlying explanatory factors hidden in the observed data and the derived learning models. Therefore, the challenge of representation learning is to have the learned latent element explanatory and disentangled from the derived abstract representation.

With the goal of discovering the underlying factors of data representation associated with particular attributes of interest, representation disentanglement is the learning task which aims at deriving a latent feature space that decomposes the derived representation so that the aforementioned attributes (e.g., face identity/pose, image style, etc.) can be identified and described. Several works have been proposed to tackle this task in unsupervised \cite{chen2016infogan,higgins2017beta}, semi-supervised \cite{cvae,makhzaniiclr14aae}, or fully supervised settings \cite{Kulkarni15dcign,acgan}. Once attribute of interest properly disentangled, one can produce the output images with particular attribute accordingly. 
\begin{figure}[t!]
	\centering
	\includegraphics[width=1.0\textwidth]{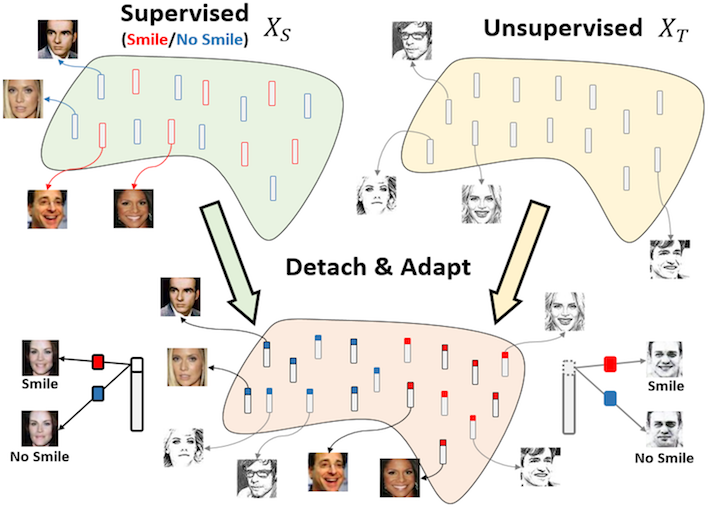}
	\vspace{-4mm}
     \caption{ 
Illustration of cross-domain representation disentanglement. With attributes observed only in the source domain, we are able to disentangle, adapt, and manipulate the data across domains with particular attributes of interest.}
	\vspace{-2mm}
	\label{fig:introduction}
\end{figure}

However, like most machine learning algorithms, representation disentanglement is not able to achieve satisfactory performances if the data to be described/manipulated are very different from the training ones. This is known as the problem of \textit{domain shift} (or \textit{domain/dataset bias}), and requires the advance of transfer learning~\cite{pan2010survey} or domain adaptation~\cite{patel2015visual} techniques to address this challenging yet practical problem. Similarly, learning of deep neural networks for interpretable and disentangled representation generally requires a large number of annotated data, and also suffers from the above problem of domain shift. 

To adapt cross-domain data for transferring the desirable knowledge such as label or attribute, one typically utilizes pre-collected or existing annotated data as source-domain training data, together with unlabeled data in the target domain of interest, for deriving the learning model. Since only unlabeled target-domain data is observed in the above scenario, it is considered as \textit{unsupervised domain adaptation}. For example, given face images with expression annotation as source-domain data ${X}_S$, and facial sketches ${X}_T$ without any annotation as target-domain data of interest (see Figure~\ref{fig:introduction} for illustration), the goal of \textit{cross-domain feature disentanglement} is to distinguish the latent feature corresponding to the expression by observing both ${X}_S$ and ${X}_T$.

\begin{figure}[t!]
	\centering
	\includegraphics[width=0.96\textwidth]{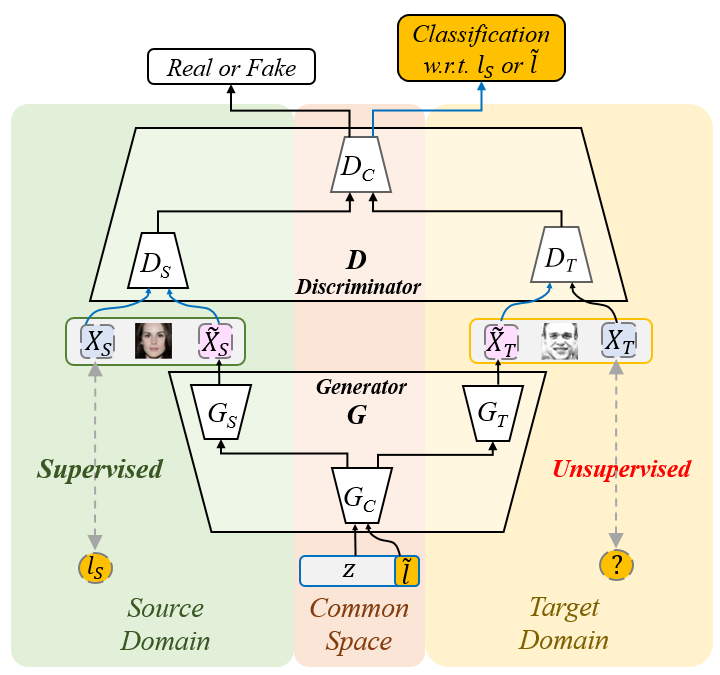}
	\vspace{-4mm}
    \caption{\small{The network architecture of Cross-Domain Representation Disentangler (CDRD). Note that while source and target-domain data are presented during training, only attribute supervision is available in the source domain, and no cross-domain data pair is needed.
}}
	\vspace{2mm}
	\label{fig:archi_wo_enc}
\end{figure}

In this paper, we propose a novel deep neural networks architecture based on generative adversarial networks (GAN)~\cite{goodfellow2014generative}. As depicted in Figure~\ref{fig:archi_wo_enc}, our proposed network observes cross-domain data with partial supervision (i.e., only annotation in ${X}_S$ is available), and performs representation learning and disentanglement in the resulting shared latent space. It is worth noting that this can be viewed as a novel learning task of joint representation disentanglement and domain adaptation in an unsupervised setting, since only unlabeled data is available in the target domain during the training stage. Later in the experiments, we will further show that the derived feature representation can be applied to describe data from both source and target domains, and classification of target-domain data can be achieved with very promising performances.

We highlight the contributions of this paper as follows:

\begin{itemize}
\item To the best of our knowledge, we are the \emph{first} to tackle the problem of representation disentanglement for cross-domain data.
\item We propose an \textit{end-to-end} learning framework for joint representation disentanglement and adaptation, while only attribute supervision is available in the source domain.
\item Our proposed model allows one to perform conditional cross-domain image synthesis and translation.
\item Our model further addresses the domain adaptation task of attribute classification. This qualitatively verifies our capability in describing and recognizing cross-domain data with particular attributes of interest.
\end{itemize}

\vspace{4mm}
\section{Related Works} \label{sec:related}

\noindent\textbf{Representation Disentanglement}

Disentangling the latent factors from the image variants has led to the understanding of the observed data~\cite{Kulkarni15dcign,acgan,cvae,makhzaniiclr14aae,chen2016infogan,higgins2017beta}. For example, by training from a sufficient amount of fully annotated data, {Kulkarni \textit{et al.}}~\cite{Kulkarni15dcign} proposed to learn interpretable and invertible graphics code when rendering image 3D models. {Odena \textit{et al.}}~\cite{acgan} augmented the architecture of generative adversarial networks (GAN) with an auxiliary classifier. Given ground truth label/attribute information during training, the model enables the synthesized images to be conditioned on the desirable latent factors. {Kingma \textit{et al.}}~\cite{cvae} extended variational autoencoder (VAE)~\cite{kingma2014stochastic} to achieve semi-supervised learning for disentanglement. {Chen \textit{et al.}}~\cite{chen2016infogan} further tackles this task in an unsupervised manner by maximizing the mutual information between pre-specified latent factors and the rendered images; however, the semantic meanings behind the disentangled factors cannot be explicitly obtained. Despite the promising progress in the above methods on deep representation disentanglement, most existing works only focus on handling and manipulating data from a single domain of interest. In practical scenarios, such settings might not be of sufficient use. This is the reason why, in this work, we aim at learning and disentangling representation across data domains in an unsupervised setting (i.e., only source-domain data are with ground truth annotation).\\

\vspace{-2mm}
\noindent\textbf{Adaptation Across Visual Domains}

Domain adaptation addresses the same learning task from data across domains. It requires one to transfer information from one (or multiple) domain(s) to another, while the domain shift is expected. In particular, unsupervised domain adaptation (UDA) deals with the task that no label supervision is available during training in the target domain. For existing UDA works, {Long \textit{et al.}}~\cite{long2013transfer} learned cross-domain projection for mapping data across domains into a common subspace. {Long \textit{et al.}}~\cite{long2014transfer} further proposed to reweight the instances across domains to alleviate the domain bias, and {Ghifary \textit{et al.}}~\cite{ghifary2017scatter} presented scatter component analysis to maximize the separability of classes and minimize the mismatch across domains. Zhang {\textit{et al.}}~\cite{Zhang_2017_CVPR} utilized coupled dimension reduction across data domains to reduce the geometrical and distribution differences.

Inspired by the adversarial learning scheme \cite{goodfellow2014generative}, several deep learning based methods have been proposed for solving domain adaptation tasks. For example, {Ganin \textit{et al.}}~\cite{ganin2015unsupervised} introduced a domain classifier in a standard architecture of convolutional neural networks (CNN), with its gradient reversal layer serving as a domain-adaptive feature extractor despite the absence of labeled data in the target domain. Similarly, {Tzeng \textit{et al.}}~\cite{tzeng2015simultaneous} utilized the domain confusion loss to learn shared representation for describing cross-domain data. {Tzeng \textit{et al.}}~\cite{Adda_CVPR2017} applied the architecture of weight sharing layers between feature extractors of source and target domains, which allows the learning of domain-specific feature embedding by utilizing such domain adversarial training strategies.

With the goal of converting images in one style to another, image--to--image translation can be viewed as another learning task that handles cross-domain visual data. For example, {Isola \textit{et al.}}~\cite{pix2pix} approached this task by applying pairs of images for learning GAN-based models. Taigman \textit{et al.}~\cite{taigman2017unsupervised} performed such tasks by employing feature consistency across domains.Without the need to observe cross-domain image pairs, Zhu \textit{et al.}~\cite{CycleGAN2017} learned the dual domain mappings with a cycle consistency loss. Similar ideas can be found in {~\cite{kim2017learning} and ~\cite{yi2017dualgan}}. Coupled GAN (CoGAN)~\cite{liu2016coupled} ties high-level information between two image domains for simultaneously rendering corresponding cross-domain images, and UNIT~\cite{liu2017unsupervised} is considered as an extended version of CoGAN, which integrates VAE and GAN to learn image translation in an unsupervised manner.

It is worth pointing out that, although approaches based on image translation are able to convert images from one domain to another, they do not exhibit the ability in learning and disentangling desirable latent representation (as ours does). As verified later in the experiments, the latent representation derived by image translation models cannot produce satisfactory classification performance for domain adaptation either.

\section{Proposed Method}\label{sec:method}

The objective of our proposed model,~\textit{Cross-Domain Representation Disentangler (CDRD)}, is to perform joint representation disentanglement and domain adaptation (as depicted in Figure~\ref{fig:archi_wo_enc}). With only label supervision available in the source domain, our CDRD derives deep \textit{disentangled feature representation} $z$ with a corresponding \textit{disentangled latent factor} $\tilde{l}$ for describing cross-domain data and their attributes, respectively. We now detail our proposed architecture of CDRD in the following subsections.

\subsection{Cross-Domain Representation Disentangler}\label{sec:CDD}

Since both {AC-GAN} \cite{acgan} and {InfoGAN}\cite{chen2016infogan} are known to learn interpretable feature representation using deep neural networks (in supervised and unsupervised settings, respectively), it is necessary to briefly review their architecture before introducing ours. Based on the recent success of {GAN} \cite{goodfellow2014generative}, both AC-GAN and InfoGAN take noise and additional class/condition as the inputs to the generator, while the label prediction is additionally performed at the discriminator for the purpose of learning disentangled features. As noted above, since both AC-GAN and InfoGAN are not designed to learn/disentangle representation for data across different domains, they cannot be directly applied for \textbf{cross-domain} representation disentanglement.

To address this problem, we propose a novel network architecture of cross-domain representation disentangler (CDRD). As depicted in Figure~\ref{fig:archi_wo_enc}, our CDRD model consists of two major components: Generators $\lbrace G_{S}, G_{T}, G_{C} \rbrace$, and Discriminators $\lbrace D_{S}, D_{T}, D_{C} \rbrace$. Similar to AC-GAN and InfoGAN, we have an auxiliary classifier attached at the end of the network, which shares all the convolutional layers with the discriminator $D_{C}$, followed by a fully connected layer to predict the label/attribute outputs. Thus, we regard our discriminator as a multi-task learning model, which not only distinguishes between synthesized and real images but also recognizes the associated image attributes. 

To handle cross-domain data with only supervision from the source domain, we choose to share weights in higher layers in $G$ and $D$, aiming at bridging the gap between high/coarse-level representations of cross-domain data. To be more precise, we split $G$ and $D$ in CDRD into multiple sub-networks specialized for describing data in the source domain $\lbrace G_{S}, D_{S}\rbrace$, target domain $\lbrace G_{T}, D_{T}\rbrace$, and the common latent space $\lbrace G_{C}, D_{C}\rbrace$ (see the green, yellow, and red-shaded colors in Figure~\ref{fig:archi_wo_enc}, respectively).

Following the challenging setting of unsupervised domain adaptation, each input image $X_{S}$ in the source domain is associated with a ground truth label $l_{S}$, while unsupervised learning is performed in the target domain. Thus, the common latent representation $z$ in the input of CDRD together with a randomly assigned attribute $\tilde{l}$ would be the inputs for the generator. For the synthesized images ${\tilde{X}}_{S}$ and $\tilde{X}_{T}$, we have:
\small
\begin{equation}\label{eq:sample_X}
\begin{aligned}
\tilde{X}_{S} \sim G_{S}(G_{C}(z, \tilde{l})), \tilde{X}_{T} \sim G_{T}(G_{C}(z, \tilde{l}))
\end{aligned}
\end{equation}
\normalsize

The objective functions for adversarial learning in source and target domain are now defined as follows:

\small
\begin{equation*}\label{eq:adv_S}
\begin{aligned}
\mathcal{L}_{adv}^{S} = \mathbb{E}[\log(D_C(D_S({X}_{S}))) ]+\mathbb{E}[\log(1-D_C(D_S( \tilde{X}_{S} )))],\\
\end{aligned}
\end{equation*}
\begin{equation*}\label{eq:adv_T}
\begin{aligned}
\mathcal{L}_{adv}^{T} = \mathbb{E}[\log(D_C(D_T({X}_{T}))) ]+\mathbb{E}[\log(1-D_C(D_T( \tilde{X}_{T} )))],\\
\end{aligned}
\end{equation*}

\begin{equation}\label{eq:adv}
\begin{aligned}
&\mathcal{L}_{adv} = \mathcal{L}_{adv}^{S} + \mathcal{L}_{adv}^{T}.
\end{aligned}
\end{equation}
\normalsize

Let {\small $P(l|X)$} be a probability distribution over labels/attributes $l$ calculated by the discriminator in CDRD. The objective functions for cross-domain representation disentanglement are defined below:
\small
\begin{equation*}\label{eq:dis_S}
\begin{aligned}
\mathcal{L}_{dis}^{S} = \mathbb{E}[\log P( l = \tilde{l} | \tilde{X}_S )]+\mathbb{E}[\log P( l = l_S | X_S )],
\end{aligned}
\end{equation*}
\begin{equation*}\label{eq:dis_T}
\begin{aligned}
&\mathcal{L}_{dis}^{T} = \mathbb{E}[\log P( l = \tilde{l} | \tilde{X}_T )]\qquad\qquad\qquad\qquad\quad\ \;\,,\\
\end{aligned}
\end{equation*}
\begin{equation}\label{eq:dis}
\begin{aligned}
\mathcal{L}_{dis} = \mathcal{L}_{dis}^{S} + \mathcal{L}_{dis}^{T}.
\end{aligned}
\end{equation}
\normalsize

\begin{algorithm}[t]
\small
\KwData{Source domain: $X_{S}$ and $l_{S}$; Target domain: $X_{T}$
}
\KwResult{Configurations of CDRD}
$\theta_{G}$, $\theta_{D} \leftarrow$ initialize\\
  \For{ Iters. of whole model}{
    $z \leftarrow$ sample from $\mathcal{N}(\textbf{0},\textbf{\textit{I}})$\\
    $\tilde{l} \leftarrow$ sample from attribute space\\
    ${\tilde{X}}_{S}$, ${\tilde{X}}_{T} \leftarrow$ sample from (\ref{eq:sample_X}) \\
    $X_{S}$, $X_{T}\leftarrow$ sample mini-batch\\
    $\mathcal{L}_{adv}$, $\mathcal{L}_{dis} \leftarrow$ calculate by (\ref{eq:adv}), (\ref{eq:dis})\\
    \For{  Iters. of updating generator }{
    $\theta_{G} \xleftarrow{+} -{\Delta}_{\theta_{G}}(-\mathcal{L}_{adv}+ \lambda \mathcal{L}_{dis})$\\
  	}
    \For{Iters. of updating discriminator }{
    $\theta_{D} \xleftarrow{+} -{\Delta}_{\theta_{D}}(\mathcal{L}_{adv}+ \lambda \mathcal{L}_{dis})$\\
  	}
  }
\caption{Learning of CDRD}\label{alg:cdrd}
\normalsize
\end{algorithm}

With the above loss terms determined, we learn our CDRD by alternatively updating Generator and Discriminator with the following gradients:
\small
\begin{equation}\label{eq:gradient}
\begin{aligned}
\theta_{G} &\xleftarrow{+} -{\Delta}_{\theta_{G}}(-\mathcal{L}_{adv}+ \lambda \mathcal{L}_{dis})\\
\theta_{D} &\xleftarrow{+} -{\Delta}_{\theta_{D}}(\mathcal{L}_{adv}+ \lambda \mathcal{L}_{dis})
\end{aligned}
\end{equation}
\normalsize
We note that the hyperparameter $\lambda$ is used to control the disentanglement ability. We will show its effect on the resulting performances in the experiments.

Similar to the concept in {InfoGAN}~\cite{chen2016infogan}, the auxiliary classifier in $D_{C}$ is to maximize the mutual information between the assigned label $\tilde{l}$ and the synthesized images in the source and target domains (i.e., $G_{S}(G_{C}(z, \tilde{l}))$ and $G_{T}(G_{C}(z, \tilde{l}))$). With network weights in high-level layers shared between source and target domains in both $G$ and $D$, the disentanglement ability is introduced to the target domain by updating the parameters in $G_{T}$ according to $\mathcal{L}_{dis}^{T}$ during the training process.

\subsection{Extended CDRD (E-CDRD)}\label{sec:CDD_ex}

\begin{figure}[t!]
	\centering
	\includegraphics[width=0.96\textwidth]{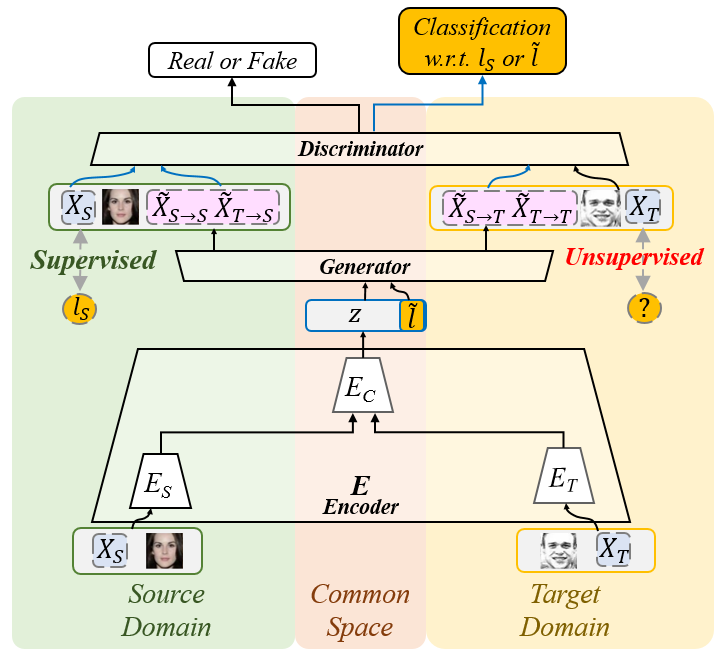}
	\vspace{-4mm}
    \caption{\small{Our proposed architecture of Extended Cross-Domain Representation Disentangler (E-CDRD), which jointly performs cross-domain representation disentanglement and image translation.}} 
	\vspace{2mm}
	\label{fig:archi_w_enc}
\end{figure}

Our CDRD can be further extended to perform joint image translation and disentanglement by adding an additional component of Encoder $\lbrace E_{S}, E_{T}, E_{C} \rbrace$ prior to the architecture of CDRD, as shown in Figure~\ref{fig:archi_w_enc}. Such Encoder-Generator pairs can be viewed as {VAE} models~\cite{kingma2014stochastic} for directly handling image variants in accordance with $\tilde{l}$.

It is worth noting that, as depicted in Figure~\ref{fig:archi_w_enc}, the Encoder $\lbrace E_{S}, E_{C} \rbrace$ and the Generator $\lbrace G_{S}, G_{C} \rbrace$ constitute a VAE module for describing source-domain data. Similar remarks can be applied for $\lbrace E_{T}, E_{C} \rbrace$ and $\lbrace G_{T}, G_{C} \rbrace$ in the target domain. It can be seen that, the components $E_{S}$ and $E_{T}$ first transform input real images $X_{S}$ and $X_{T}$ into a common feature, which is then encoded by $E_{C}$ as latent representation:
\small
\begin{equation}\label{eq:sample_z}
\begin{aligned}
z_S &\sim E_C(E_S(X_S)) = q_{S}(z_S|X_S), \\
z_T &\sim E_C(E_T(X_T)) = q_{T}(z_T|X_T). \\
\end{aligned}
\end{equation}
\normalsize

Once the latent representations $z_S$ and $z_T$ are obtained, the remaining architecture is the standard CDRD, which can be applied to recover the images with the assigned $\tilde{l}$ in the associated domains, i.e. $\tilde{X}_{S \rightarrow S}$ and $\tilde{X}_{T \rightarrow T}$:
\small
\begin{equation}\label{eq:sample_X_recon}
\begin{aligned}
\tilde{X}_{S \rightarrow S} \sim G_S(G_C(z_{S}, \tilde{l})), \tilde{X}_{T \rightarrow T} \sim G_T(G_C(z_{T}, \tilde{l})).
\end{aligned}
\end{equation}
\normalsize

\begin{algorithm}[ht]
\small
\KwData{Source domain: $X_{S}$ and $l_{S}$; Target domain: $X_{T}$
}
\KwResult{Configurations of E-CDRD}
$\theta_{E}$, $\theta_{G}$, $\theta_{D} \leftarrow$ initialize\\
  \For{ Iters. of whole model}{
    $z \leftarrow$ sample from $\mathcal{N}(\textbf{0},\textbf{\textit{I}})$\\
    $X_{S}$, $X_{T}\leftarrow$ sample mini-batch\\
    $z_{S}$, $z_{T}\leftarrow$ sample from (\ref{eq:sample_z})\\
    $\tilde{l} \leftarrow$ sample from attribute space\\
    ${\tilde{X}}_{S}$, ${\tilde{X}}_{T} \leftarrow$ sample from (\ref{eq:sample_X}) \\
    ${\tilde{X}}_{S \rightarrow S}$, ${\tilde{X}}_{T \rightarrow T}$, ${\tilde{X}}_{S \rightarrow T}$, ${\tilde{X}}_{T \rightarrow S} \leftarrow$ sample from (\ref{eq:sample_X_recon}), (\ref{eq:sample_X_cross}) \\
    $\mathcal{L}_{vae}$, $\mathcal{L}_{adv}$, $\mathcal{L}_{dis} \leftarrow$ calculate by (\ref{eq:vae}), (\ref{eq:adv_ex}), (\ref{eq:dis_ex})\\
    \For{  Iters. of updating encoder }{
    $\theta_{E} \xleftarrow{+} -{\Delta}_{\theta_{E}}(\mathcal{L}_{vae})$\\
  	}
    \For{  Iters. of updating generator }{
    $\theta_{G} \xleftarrow{+} -{\Delta}_{\theta_{G}}(\mathcal{L}_{vae}-\mathcal{L}_{adv}+ \lambda \mathcal{L}_{dis})$\\
  	}
    \For{Iters. of updating discriminator }{
    $\theta_{D} \xleftarrow{+} -{\Delta}_{\theta_{D}}(\mathcal{L}_{adv}+ \lambda \mathcal{L}_{dis})$\\
  	}
  }
\caption{Learning of E-CDRD}\label{alg:e-cdrd}
\normalsize
\end{algorithm}

The VAE regularizes the Encoder by imposing a prior over the latent distribution $p(z)$. Typically we have $z \sim \mathcal{N}(\textbf{0},\textbf{\textit{I}})$. In E-CDRD, we advance the objective functions of VAE for each data domain as follows:

\small
\begin{equation*}\label{eq:vae_S}
\begin{aligned}
\mathcal{L}_{vae}^{S} = {\lVert\Phi(X_{S}) - \Phi(\tilde{X}_{S\rightarrow S})\rVert}_{F}^{2}+\textit{KL}(q_{S}(z_S|X_S)||p(z))
\end{aligned}
\end{equation*}
\begin{equation*}\label{eq:vae_T}
\begin{aligned}
\mathcal{L}_{vae}^{T} = {\lVert\Phi(X_{T}) - \Phi(\tilde{X}_{T \rightarrow T})\rVert}_{F}^{2}+\textit{KL}(q_{T}(z_T|X_T)||p(z))
\end{aligned}
\end{equation*}

\begin{equation}\label{eq:vae}
\begin{aligned}
\mathcal{L}_{vae} = \mathcal{L}_{vae}^{S} + \mathcal{L}_{vae}^{T}
\end{aligned}
\end{equation}
\normalsize
\noindent where the first term denotes the \textit{perceptual loss}~\cite{johnson2016perceptual}, which calculates the reconstruction error between the synthesized output $\tilde{X}_{S\rightarrow S}$ (or $\tilde{X}_{T\rightarrow T}$) and its original input ${X}_S$ (or ${X}_T$) with network transformation $\Phi$ (the similarity metric of~\cite{larsen2016autoencoding} is applied in our work). On the other hand, the second term indicates \textit{Kullback-Leibler divergence} over the auxiliary distribution $q_{S}(z_S|X_S)$, $q_{T}(z_T|X_T)$ and the prior $p(z)$. 

Moreover, similar to the task of image translation, our E-CDRD also outputs images in particular domains accordingly, i.e. $\tilde{X}_{S \rightarrow T}$ is translated from the source to target domain, and $\tilde{X}_{T \rightarrow S}$ is the output from the target to source domain:
\small
\begin{equation}\label{eq:sample_X_cross}
\begin{aligned}
\tilde{X}_{S \rightarrow T} \sim G_T(G_C(z_{S}, \tilde(l))), \tilde{X}_{T \rightarrow S} \sim G_S(G_C(z_{T}, \tilde(l))).
\end{aligned}
\end{equation}
\normalsize

With the above observations, the objective functions with adversarial learning for E-CDRD are modified as follows:
\small
\begin{equation*}\label{eq:adv_S_ex}
\begin{aligned}
\mathcal{L}_{adv}^{S} = \mathbb{E}[\log(D_C(D_S({X}_{S}))) ]&+\mathbb{E}[\log(1-D_C(D_S( \tilde{X}_{S} )))]\\
&+\mathbb{E}[\log(1-D_C(D_S( \tilde{X}_{S \rightarrow S} )))]\\
&+\mathbb{E}[ \log(1-D_C(D_S( \tilde{X}_{T \rightarrow S} )))],
\end{aligned}
\end{equation*}
\begin{equation*}\label{eq:adv_T_ex}
\begin{aligned}
\mathcal{L}_{adv}^{T} = \mathbb{E}[\log(D_C(D_T({X}_{T}))) ]&+\mathbb{E}[\log(1-D_C(D_T( \tilde{X}_{T} )))]\\
& +\mathbb{E}[\log(1-D_C(D_T( \tilde{X}_{T \rightarrow T} )))]\\
& +\mathbb{E}[\log(1-D_C(D_T( \tilde{X}_{S \rightarrow T} )))],
\end{aligned}
\end{equation*}
\begin{equation}\label{eq:adv_ex}
\begin{aligned}
\mathcal{L}_{adv} = \mathcal{L}_{adv}^{S} + \mathcal{L}_{adv}^{T}.
\end{aligned}
\end{equation}
\normalsize
Similarly, we revise the objective functions for representation disentanglement as follows:
\small
\begin{equation*}\label{eq:dis_S_ex}
\begin{aligned}
\mathcal{L}_{dis}^{S} &= \mathbb{E}[\log P( l = \tilde{l} | \tilde{X}_S )]+\mathbb{E}[\log P( l = l_S | X_S )] \\
&+\mathbb{E}[\log P( l = \tilde{l} | \tilde{X}_{S\rightarrow S} )] 
+\mathbb{E}[\log P( l = \tilde{l} | \tilde{X}_{T\rightarrow S} )]
\end{aligned}
\end{equation*}
\begin{equation*}\label{eq:dis_T_ex}
\begin{aligned}
\mathcal{L}_{dis}^{T} &= \mathbb{E}[\log P( l = \tilde{l} | \tilde{X}_T )] \\
&+\mathbb{E}[\log P( l = \tilde{l} | \tilde{X}_{T\rightarrow T} )] 
+\mathbb{E}[\log P( l = \tilde{l} | \tilde{X}_{S\rightarrow T} )]
\end{aligned}
\end{equation*}
\begin{equation}\label{eq:dis_ex}
\begin{aligned}
\mathcal{L}_{dis} = \mathcal{L}_{dis}^{S} + \mathcal{L}_{dis}^{T}.
\end{aligned}
\end{equation}
\normalsize

To train our E-CDRD, we alternatively update Encoder, Generator and Discriminator with the following gradients:
\small
\begin{equation}
\begin{aligned}
\theta_{E} &\xleftarrow{+} -{\Delta}_{\theta_{E}}(\mathcal{L}_{vae})\\
\theta_{G} &\xleftarrow{+} -{\Delta}_{\theta_{G}}(\mathcal{L}_{vae}-\mathcal{L}_{adv}+ \lambda \mathcal{L}_{dis})\\
\theta_{D} &\xleftarrow{+} -{\Delta}_{\theta_{D}}(\mathcal{L}_{adv}+ \lambda \mathcal{L}_{dis})
\end{aligned}
\end{equation}
\normalsize
It is worth noting that, by jointly considering the above objective functions of VAE and those for adversarial and disentanglement learning, our E-CDRD can be applied for conditional cross-domain image synthesis and translation. Similar to CDRD, the hyperparameter $\lambda$ controls the ability of E-CDRD for performing disentanglement (and will be analyzed in the experiments). 

Finally, the pseudo code for training our CDRD and E-CDRD are summarized in Algorithms~\ref{alg:cdrd} and~\ref{alg:e-cdrd}, respectively. Implementation details of our network architectures will be presented in the supplementary materials.

\section{Experiments} \label{sec:results}
{
We now evaluate the performance of our proposed method, which is applied to perform cross-domain representation disentanglement and adaptation simultaneously. As noted in Section~\ref{sec:CDD}, the discriminator in our CDRD (or E-CDRD) is augmented with an auxiliary classifier, which classifies images with respect to the disentangled latent factor $l$. With only supervision from the source-domain data, such a classification task is also considered as the task of unsupervised domain adaptation (UDA) for cross-domain visual classification. We will also provide quantitative UDA results to further support the use of our model for describing, manipulating, and recognizing cross-domain data with particular attributes of interest.

\subsection{Datasets}
{

\begin{figure}[t!]
	\vspace{-4mm}
    \centering
	\includegraphics[width=1\textwidth]{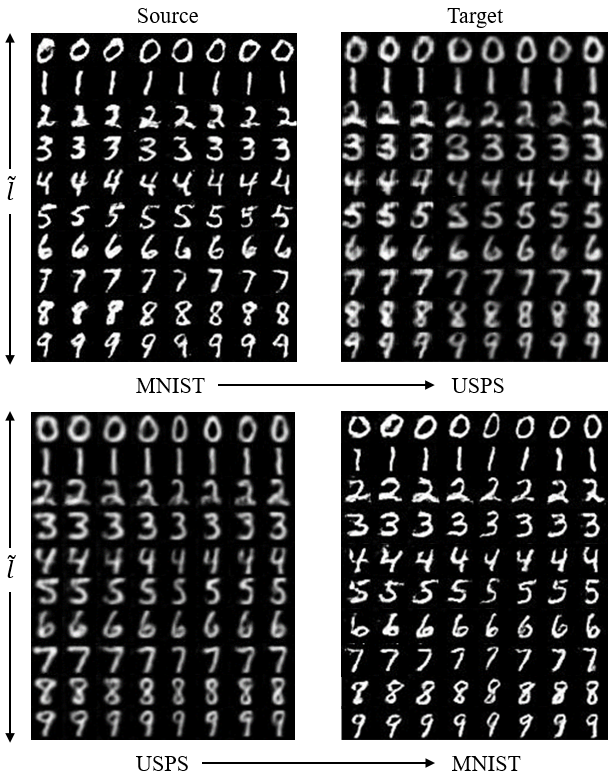}
	\vspace{-4mm}
    \caption{
    {
    \small{Cross-domain conditional image synthesis for MNIST $\rightarrow$ USPS and USPS $\rightarrow$ MNIST with the attribute $\tilde{l}$ as \textit{digits}.}
}
    }
    
	\label{fig:digit_syn_2}
\end{figure}
We consider three different types of datasets, including digit, face, and scene, for performance evaluation:

\textbf{Digits.} \textit{MNIST}, \textit{USPS} and \textit{Semeion} \cite{Lichman:2013} are hand-written digit image datasets, which are viewed as different data domains. MNIST contains 60K/10K instances for training/testing, and USPS consists of 7291/2007 instances for training/testing. Semeion contains 1593 handwritten digits provided by about 80 persons, stretched in a rectangular box 16x16 in a gray scale of 256 values. We resize these images to 28x28 pixels to match the resolution of the images in MNIST and USPS. For UDA, we follow the same protocol in~\cite{long2013transfer,long2014transfer} to construct source and target-domain data with the digits as the attributes.

\textbf{Faces.} We consider facial \textit{photo} and \textit{sketch} images as data in different domains. For facial photo images, we consider the CelebFaces Attributes dataset (CelebA)~\cite{liu2015faceattributes}, which is a large-scale face image dataset including more than $200K$ celebrity photos annotated with $40$ facial attributes. We randomly select half of the dataset as the photo domain, then convert the other half into sketch images based on the procedure used in \cite{pix2pix} (which thus results in our sketch domain data). For simplicity, among the $40$ attributes of face images, we choose \textit{``glasses''} and \textit{``smiling''} as the attributes of interest. The common rule of thumb 80/20 is used for the training/testing dataset split.

\textbf{Scenes.} We have \textit{photo} and \textit{paint} images as scene image data in different domains. We collect 1,098 scene photos from Google Image Search and Flickr. We randomly select half of the photo collection as the photo domain and apply the style transfer method in \cite{CycleGAN2017} on the rest half to produce the painting images. 
Each image is manually labeled as \textit{``night''}, i.e. day/night, and \textit{``season''}, i.e. winter/summer, for attribute of interest. We use 80\% of all the data in each domain for training and the rest 20\% for testing. 

It is worth repeating that, while the image data in both domains are presented during training, we do not require any paired cross-domain image pairs to learn our models (neither do~\cite{CycleGAN2017,liu2016coupled,liu2017unsupervised}). And, for fair comparisons, the ground truth attribute is only available for the source-domain data for all the methods considered.

\vspace{3mm}

\vspace{-2mm}
\subsection{Cross-Domain Representation Disentanglement and Translation}\label{ssec:result1}
\begin{figure}[t!]
	\vspace{-2mm}
    \centering
	\includegraphics[width=1\textwidth]{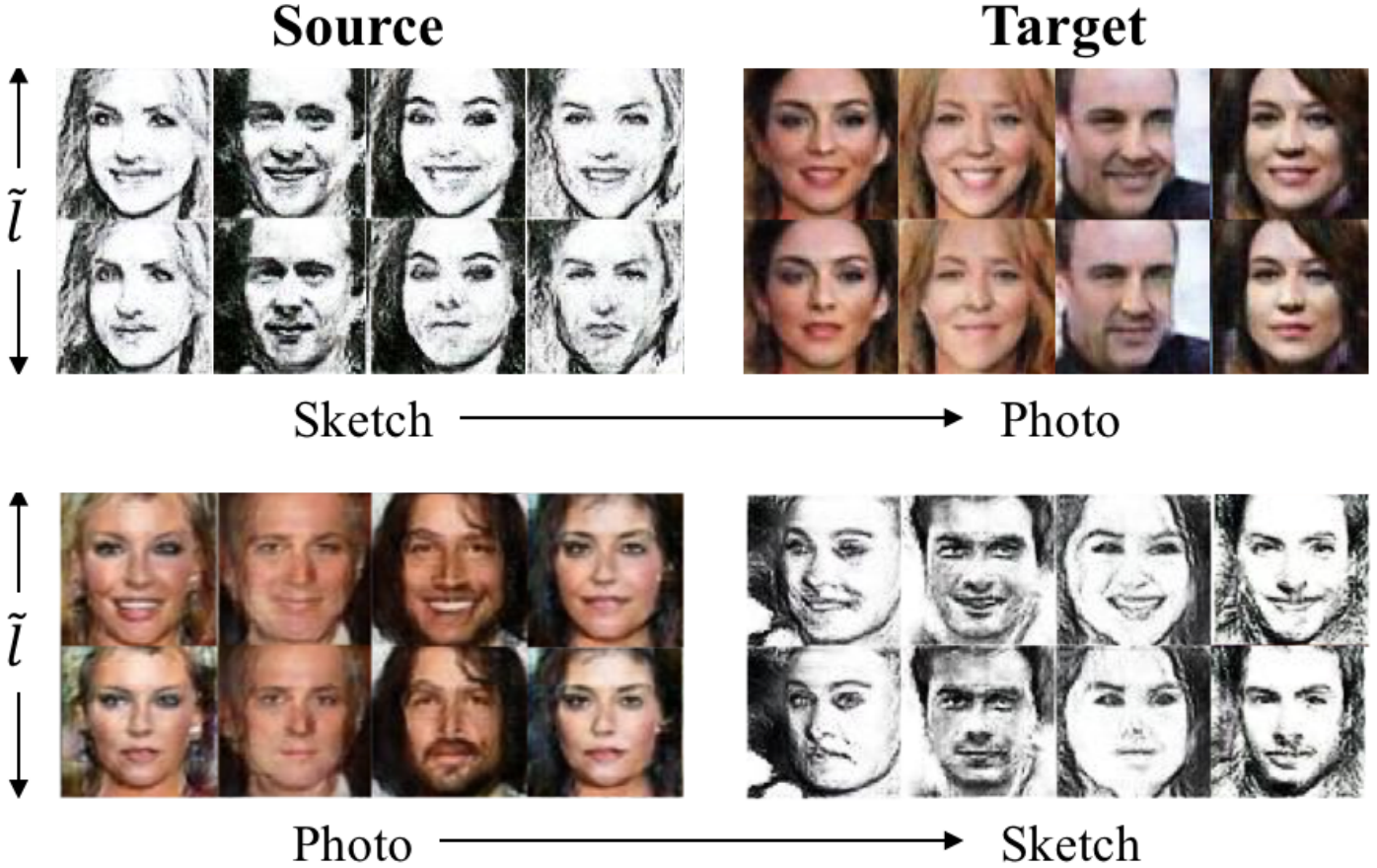}
	\vspace{-4mm}
    \caption{
    {
    \small{Cross-domain conditional image synthesis for Sketch $\rightarrow$ Photo and Photo $\rightarrow$ Sketch with $\tilde{l}$ as \textit{smiling}. Note that the identities are different across image domains.}
}
    }
    
	\label{fig:face_transfer}
\end{figure}

We first conduct \textit{conditional image synthesis} to evaluate the effectiveness of CDRD for representation disentanglement. Recall that the architecture of our CDRD allows one to freely control the disentangled factor $\tilde{l}$ via~\eqref{eq:sample_X} with randomly sampled $z$ to produce the corresponding output.

\textbf{Single Source Domain vs. Single Target Domain.} 
Considering a pair of data domains from each image type (i.e., digit, face, or scene images), we plot the results of conditional image synthesis in Figures~\ref{fig:digit_syn_2} and \ref{fig:face_transfer}. From these results, we have a random vector $z$ as the input in each column, and verify that the images at either domain can be properly synthesized and manipulated (based on the attribute of interest).

\textbf{Single Source vs. Multiple Target Domains.}
We now extend our CDRD to perform cross-domain representation disentanglement, in which a single source domain and multiple target domains are of use. From the results shown in Figure~\ref{fig:multi_dms_digit}, we see that our CDRD can be successfully applied for this challenging task even with only attribute supervision from the single source-domain data. This confirms our design of high-level sharing weights in CDRD.

\begin{figure}[t!]
	\vspace{-2mm}
	\centering
	\includegraphics[width=1\textwidth]{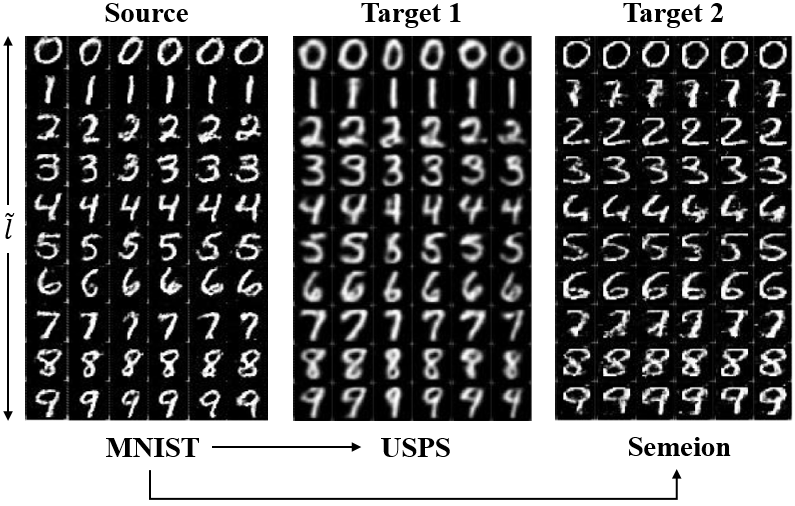}
	\vspace{-4mm}
    \caption{
    {
    \small{Cross-domain conditional image synthesis from a single source to multiple target domains: MNIST $\rightarrow$ USPS and Semeion with $\tilde{l}$ as \textit{digits}.}
}
    }
    
	\label{fig:multi_dms_digit}
\end{figure}

Next, we evaluate our performance for \textit{conditional image--to--image translation}. This requires the use of our E-CDRD for joint cross-domain representation disentanglement and translation, i.e., a representation encoded from one domain can be translated to another with the specified attribute value $\tilde{l}$.
\begin{figure}[t!]
	\vspace{-2mm}
    \centering
	\includegraphics[width=1\textwidth]{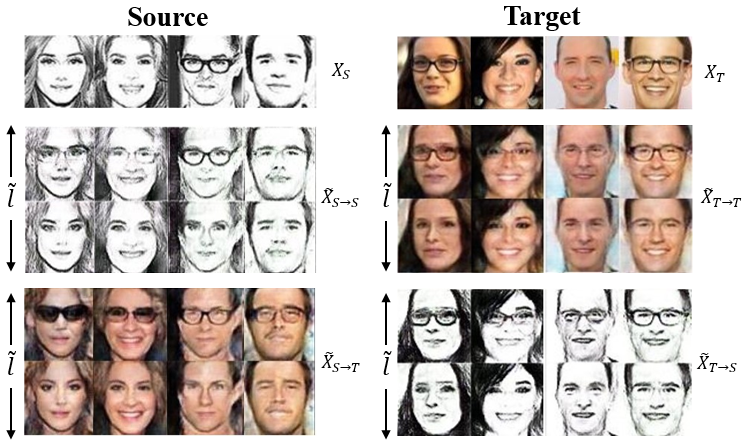}
	\vspace{-4mm}
    \caption{
    {
    \small{Cross-domain conditional image translation for facial Sketch $\rightarrow$ Photo with $\tilde{l}$ as \textit{glasses}.}
}
    }
    
	\label{fig:face_recon_trans}
\end{figure}
\begin{figure}[t!]
	\vspace{-2mm}
    \centering
	\includegraphics[width=1\textwidth]{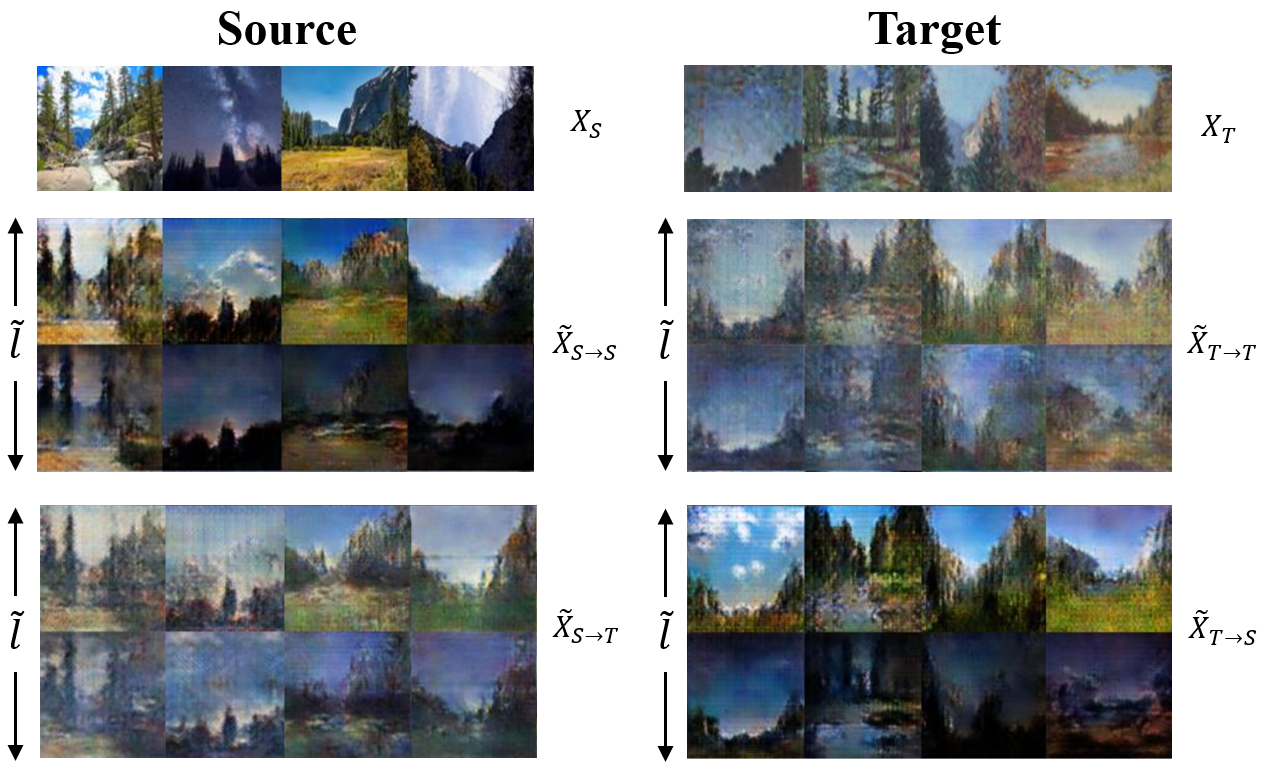}
	\vspace{-4mm}
    \caption{
    {
    \small{Cross-domain conditional image translation for scene images: Photo $\rightarrow$ Paint with $\tilde{l}$ as \textit{night}.}
}
    }
	\label{fig:scene_recon_trans}
\end{figure}

We utilize face and scene image data, as shown in Figures~\ref{fig:face_recon_trans} and~\ref{fig:scene_recon_trans}, respectively. Take Figure~\ref{fig:face_recon_trans} as example, when having a facial sketch as the input in source domain, our E-CDRD is able to manipulate the attribute of glasses not only for the output image in the same domain (i.e., sketch), but also for the facial photo in the target domain (e.g., photo). 

The above results support the use of our CDRD and E-CDRD for learning disentangled feature representation from cross-domain data, and confirm its effectiveness in producing or translating images with manipulated attributes in either domain.

As an additional remark, for conditional image translation, one might consider an alternative solution, which first performs conditional image synthesis using source-domain data, followed by using existing off-the-shelf image-to-image translation frameworks to convert such outputs into the images in the target domain. However, such integrated approaches cannot guarantee that proper disentangled representation can be obtained in the shared feature space. 
}

\subsection{Unsupervised Domain Adaptation}\label{ssec:result2}

\begin{table*}[]
\centering
\caption{\small UDA accuracy (\%) for recognizing target-domain images with the attribute of digits (0-9). Take M $\rightarrow$ U for example, we set MNIST and USPS as source and target domains, respectively.}
\vspace{-2mm}
\label{tab:dig}
\begin{tabular}{cccccccccccccc}
		& \small \begin{tabular}[c]{@{}c@{}}GFK\\ \cite{gong2012geodesic}\end{tabular}   & \small \begin{tabular}[c]{@{}c@{}}JDA\\ \cite{long2013transfer}\end{tabular}   & \small \begin{tabular}[c]{@{}c@{}}SA\\ \cite{fernando2013unsupervised}\end{tabular} & \small \begin{tabular}[c]{@{}c@{}}TJM\\ \cite{long2014transfer}\end{tabular}   & \small \begin{tabular}[c]{@{}c@{}}SCA\\ \cite{ghifary2017scatter}\end{tabular}   & \small \begin{tabular}[c]{@{}c@{}}JGSA\\ \cite{Zhang_2017_CVPR}\end{tabular}  & \small \begin{tabular}[c]{@{}c@{}}DC\\ \cite{tzeng2015simultaneous}\end{tabular}    & \small \begin{tabular}[c]{@{}c@{}}GR\\ \cite{ganin2015unsupervised}\end{tabular}    & \small \begin{tabular}[c]{@{}c@{}}CoGAN\\ \cite{liu2016coupled}\end{tabular} & \small \begin{tabular}[c]{@{}c@{}}ADDA\\ \cite{Adda_CVPR2017}\end{tabular}  & \small \begin{tabular}[c]{@{}c@{}}DRCN\\ \cite{ghifary2016deep}\end{tabular}     & \small \begin{tabular}[c]{@{}c@{}}ADGAN\\ \cite{sankaranarayanan2017generate}\end{tabular}      & \small \begin{tabular}[c]{@{}c@{}}CDRD\\ \  \end{tabular}  \\   \cline{1-14} \hline
\small M $\rightarrow$ U  & \small 67.22 & \small 67.28 & \small 67.78 & \small 63.28 & \small 65.11 & \small 80.44 & \small 79.10 & \small 77.10 & \small 91.20 & \small 89.40 & \small 91.80 & \small 92.50 & \small \textbf{95.05} \\
\small U $\rightarrow$ M  & \small 46.45 & \small 59.65 & \small 48.80 & \small 52.25 & \small 48.00 & \small 68.15 & \small 66.50 & \small 73.00 & \small 89.10 & \small 90.10 & \small 73.67 & \small 90.80 & \small \textbf{94.35} \\  \cline{1-14}  
\small Average & \small 56.84 & \small 63.47 & \small 58.29 & \small 57.77 & \small 56.55 & \small 74.30 & \small 72.80 & \small 75.05 & \small 90.15 & \small 89.75 & \small 82.74 & \small 91.65 & \small \textbf{94.70}
\end{tabular}
\end{table*}

Finally, we verify that our model can be applied to image classification, i.e., use of the discriminator in our model for recognizing images with particular attribute $l$. As noted above, this can be viewed as the task of UDA since only supervision is available in the source domain during training.
 
\textbf{Digits.} 
For UDA with digit images, we consider MNIST$\rightarrow$USPS and USPS$\rightarrow$MNIST, and we evaluate the classification accuracy for target-domain images. Table~\ref{tab:dig} lists and compares the performances of recent UDA methods. We can see that a significant improvement was achieved by our CDRD. 

It is worth noting that, while UNIT~\cite{liu2017unsupervised} reported 0.9597 for M$\rightarrow$U and 0.9358 for U$\rightarrow$M, UPDAG~\cite{Bousmalis_2017_CVPR} achieved 0.9590 for M$\rightarrow$U, they considered much larger datasets (UNIT required 60000/7291 images for MNIST/USPS, and UPDAG required 50000/6562 for MNIST/USPS). We follow the same protocol in~\cite{long2013transfer,long2014transfer} for reporting our results in Table~\ref{tab:dig}.

\begin{figure}[t!]
    \centering
    \begin{subfigure}[t]{0.5\textwidth}
        \centering
        \includegraphics[width=0.85\textwidth]{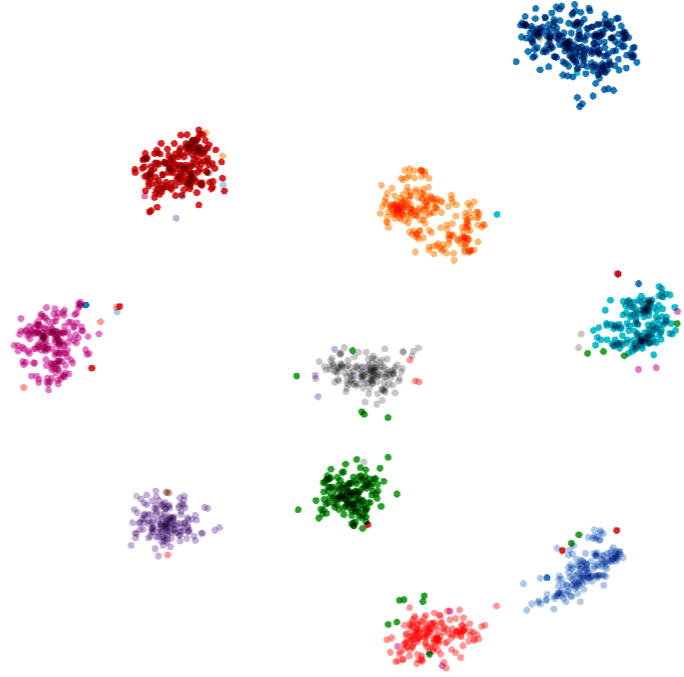}
        \caption{Colorize w.r.t. \textbf{class 0-9}.}
        \label{fig:digit_tsne1}
    \end{subfigure}%
    ~ 
    \begin{subfigure}[t]{0.5\textwidth}
        \centering
        \includegraphics[width=0.85\textwidth]{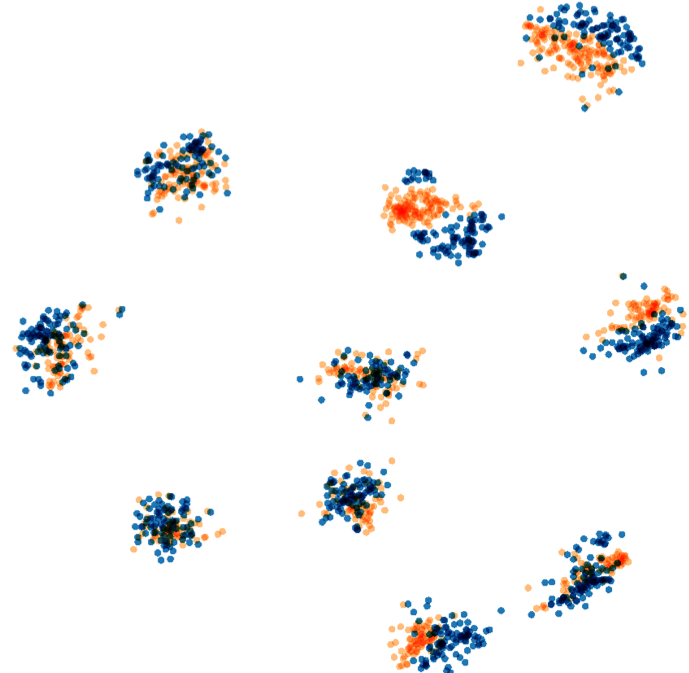}
        \caption{Colorize w.r.t. \textbf{source/target}.}
        \label{fig:digit_tsne2}
    \end{subfigure}
    \vspace{-2mm}
    \caption{\small{t-SNE visualization of the handwritten image features for MNIST$\rightarrow$USPS. Difference colors indicate (a) attribute and (b) domain information.}}
    
    \label{fig:digit_tsne}
\end{figure}
\begin{table*}[!htp]
\centering
\caption{\small UDA accuracy (\%) of cross-domain classification on face and scene images.}
\vspace{-2mm}
\begin{subtable}{.5\textwidth}
\centering
\caption{Faces.}
\vspace{-2mm}
\label{tab:face}
\begin{tabular}{cccccc}
\small Domain    & \small $\tilde{l}$          & \small CoGAN & \small UNIT  & \small CDRD  & \small E-CDRD \\ \hline
\small sketch ($\mathcal{S}$) & \small smiling    & \small 89.50 & \small 90.10 & \small 90.19 & \small 90.01  \\
\small photo ($\mathcal{T}$)  &  -         & \small 78.90 & \small 81.04 & \small \textbf{87.61} & \small \textbf{88.28}  \\ \hline
\small sketch ($\mathcal{S}$) & \small glasses & \small 96.63 & \small 97.65 & \small 97.06 & \small 97.19  \\
\small photo ($\mathcal{T}$)  &   -         & \small 81.01 & \small 79.89 & \small \textbf{94.49} & \small \textbf{94.84} 
\end{tabular}

\end{subtable}% <---- don't forget this %
\begin{subtable}{.5\textwidth}
\centering
\caption{Scenes.}
\vspace{-2mm}
\label{tab:scene}
\begin{tabular}{cccccc}
\small Domain    & \small $\tilde{l}$          & \small CoGAN & \small UNIT  & \small CDRD  & \small E-CDRD \\ \hline
\small photo ($\mathcal{S}$) & \small night    & \small 98.04 & \small 98.49 & \small 97.06 & \small 97.14  \\
\small paint ($\mathcal{T}$)  &    -        & \small 65.18 & \small 67.81 & \small \textbf{84.21} & \small \textbf{85.58}  \\ \hline
\small photo ($\mathcal{S}$) & \small season & \small 86.74 & \small 85.64 & \small 86.21 & \small 88.92  \\
\small paint ($\mathcal{T}$)  &     -       & \small 65.94 & \small 66.09 & \small \textbf{79.87} & \small \textbf{80.03} 
\end{tabular}
\end{subtable}
\vspace{-4mm}
\end{table*}

In addition, we extract latent features from the last shared layer (prior to the auxiliary classifier) in Discriminator. We visualize such projected features via t-SNE, and show the results in Figure~\ref{fig:digit_tsne}. From Figure~\ref{fig:digit_tsne1}, we see that the image features of each class of digits were well separated, while the features of the same class but from different domains were properly clustered (see Figure~\ref{fig:digit_tsne2}). This confirms the effectiveness of our model in describing and adapting cross-domain image data.

\textbf{Faces and Scenes.} Tables~\ref{tab:face} and~\ref{tab:scene} show our UDA performance and comparisons using cross-domain face and scene images, respectively. It can be seen that, neither CoGAN nor UNIT were able to produce satisfactory performances, as the performance gaps between source and target-domain images were from about 10\% to 30\%. In contrast, the use of our E-CDRD reported much smaller performance gaps, and confirms that our model is preferable for translating and adapting cross-domain images with particular attributes of interest. 

It is worth noting that our E-CDRD reported further improved results than CDRD, since joint disentanglement and translation is performed when learning E-CDRD, which results in improved representation for describing cross-domain data. Another remark is that, by observing synthesized data with given assigned label $\tilde{l}$, our classifier is able to observe target domain data together with assigned attribute information. This is different from traditional domain adaptation methods, as our method breaks the limitation of lacking of ground truth attributes in target domain.
\subsection{Sensitivity Analysis}\label{ssec:result4}

\begin{figure}[t!]
	\vspace{0mm}
	\centering
	\includegraphics[width=0.85\textwidth]{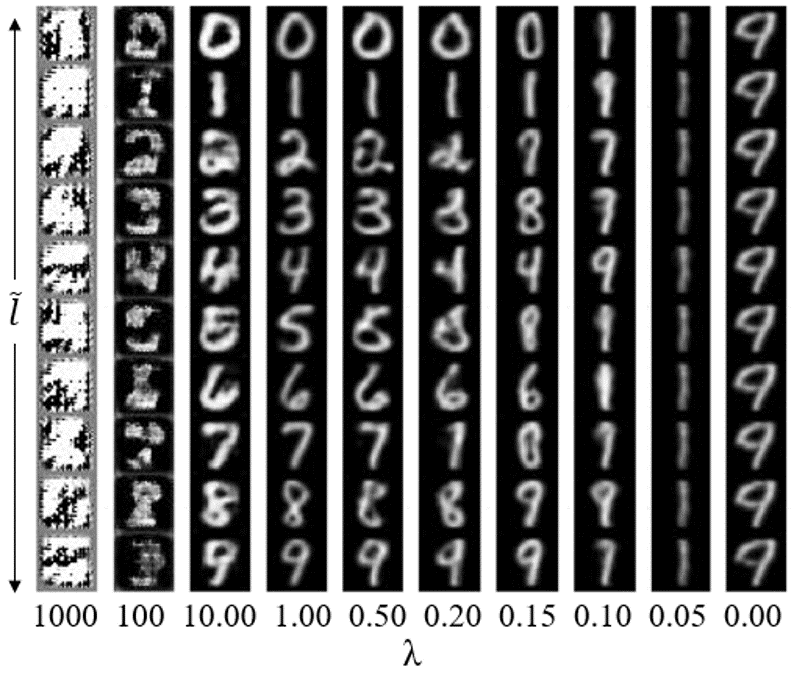}
	\vspace{-2mm}
    \caption{
    {
    \small{Sensitivity analysis on $\lambda$ for M$\rightarrow$U. Each column shows synthesized USPS images by a $\lambda$ choice, while the elements in each column are expected to be associated with $\tilde{l}=0-9$.}
}
    }
    
	\label{fig:sensitivity}
\end{figure}

As noted in Section~\ref{sec:method}, we have a hyperparameter $\lambda$ in~\eqref{eq:gradient} controlling the disentanglement ability of our model. In order to analyze its effect on the performance, we vary $\lambda$ from 0.00 to 1000 and plot the corresponding disentangled results in Figure~\ref{fig:sensitivity}. From this figure, we see that smaller $\lambda$ values were not able to manipulate images with different attributes, while extremely large $\lambda$ would result in degraded image quality (due to the negligence of the image adversarial loss). Thus, the choice of $\lambda$ between 0.5 and 10 would be preferable (we set and fix $\lambda=1$ in all our experiments).

\section{Conclusions} \label{sec:conclusions}

We presented a deep learning framework of Cross-Domain Representation Disentangler (CDRD) and its extension (E-CDRD). Our models perform joint representation disentanglement and adaption of cross-domain images, while only attribute supervision is available in the source domain. We successfully verified that our models can be applied to conditional cross-domain image synthesis, translation, and the task of unsupervised domain adaptation.\\

\noindent\textbf{Acknowledgments}
This work was supported in part by the Ministry of Science
and Technology of Taiwan under grants MOST 107-2636-E-009-001 and 107-2634-F-002-010.

%%%%%%%%%%%%%%%

{\small
\bibliographystyle{ieee}
\bibliography{egbib}
}

\end{document}